\documentclass{Interspeech}

\interspeechcameraready

\title{\textit{Neuro2Semantic}: A Transfer Learning Framework for Semantic Reconstruction of Continuous Language from Human Intracranial EEG}

\author[affiliation={1}]{Siavash}{Shams}
\author[affiliation={1}]{Richard}{Antonello}
\author[affiliation={1}]{Gavin}{Mischler}
\author[affiliation={2}]{Stephan}{Bickel}
\author[affiliation={2}]{Ashesh}{Mehta}
\author[affiliation={1}]{Nima}{Mesgarani}

\affiliation{Department of Electrical Engineering}{Columbia University}{USA}
\affiliation{}{The Feinstein Institutes for Medical Research}{USA}
\email{ss6928@columbia.edu}

\usepackage{comment}

\begin{document}

\maketitle
\renewcommand\thefootnote{}\footnotetext{To appear in Interspeech 2025}
\renewcommand\thefootnote{\arabic{footnote}} 

\begin{abstract}
    Decoding continuous language from neural signals remains a significant challenge in the intersection of neuroscience and artificial intelligence. We introduce Neuro2Semantic, a novel framework that reconstructs the semantic content of perceived speech from intracranial EEG (iEEG) recordings. Our approach consists of two phases: first, an LSTM-based adapter aligns neural signals with pre-trained text embeddings; second, a corrector module generates continuous, natural text directly from these aligned embeddings. This flexible method overcomes the limitations of previous decoding approaches and enables unconstrained text generation. Neuro2Semantic achieves strong performance with as little as 30 minutes of neural data,  outperforming a recent state-of-the-art method in low-data settings. These results highlight the potential for practical applications in brain-computer interfaces and neural decoding technologies.
\end{abstract}

\section{Introduction}

Recent advances at the intersection of artificial intelligence (AI) and neuroscience have enabled powerful new modeling capabilities, particularly in the development of neural decoding models. These models aim to reconstruct stimuli or intentions based on measured neural activity~\cite {kriegeskorte2019interpreting}. Decoding models have been explored across various neuroimaging modalities, including intracranial EEG  (iEEG)~\cite{chakrabarti2015progress, akbari2019towards, wang2023brainbert}, functional magnetic resonance imaging (fMRI)~\cite{naselaris2011encoding,  tang2023semantic}, magnetoencephalography (MEG)~\cite{defossez2022decoding, wang2024semantic}, and electroencephalography (EEG) \cite{wang2022open,liu2024eeg2text}. These models have been applied to diverse settings such as imagined and perceived language~\cite{wang2022open, defossez2022decoding, tang2023semantic}, speech reconstruction \cite{akbari2019towards, li2024neural2speech, lee24c_interspeech},  motor control~\cite{robinson2016noninvasive, pandarinath2017high}, and vision~\cite{ zou2023generalized, xia2024dream, benchetrit2024brain}. Of particular note are recent efforts showcasing the ability of these models to decode motor intention for speech at near real-time speeds with high accuracy~\cite{willett2023high, metzger2023high}. Such models have the potential to revolutionize speech therapies for those who suffer from maladies that make it difficult to produce speech. However, these approaches primarily focus on decoding motor intentions, which may not capture the full richness of linguistic semantic content.

An alternative to decoding motor intention of speech is decoding the semantics of intended speech from elsewhere in cortex~\cite{huth2016natural, rybavr2022neural}. While semantic decoding has been investigated using fMRI and MEG~\cite{tang2023semantic, dash2020decoding}, there is a less research leveraging the higher temporal resolution and signal quality of iEEG for this purpose~\cite{makin2020machine}. Despite the potential advantages of using iEEG for semantic decoding, existing methods face significant challenges when adapting to this domain, particularly due to data scarcity. To address these limitations, we propose Neuro2Semantic, a novel framework that employs transfer learning to efficiently decode the semantics of perceived speech from iEEG recordings with limited data availability.

Our approach has two main parts. First, we train an LSTM \cite{10.1162/neco.1997.9.8.1735} adapter to align neural data with a pre-trained text embedding space~\cite{raffel_exploring_2020, OpenAI2022Embeddings} using a contrastive loss function. Second, after aligning the neural embeddings, we fine-tune a pre-trained text reconstruction model~\cite{morris_text_2023} to extract coherent text from the neural-aligned embeddings. This step allows for unconstrained text generation, moving beyond classification-based methods that are restricted to predefined vocabularies or limited sets of candidates.

We demonstrate that our Neuro2Semantic framework can successfully perform few-shot reconstruction of the meaning of perceived speech in in-domain settings with as little as 30 minutes of neural data. Moreover, it achieves strong performance in zero-shot reconstruction in out-of-domain contexts, showcasing its ability to generalize to entirely new semantic content. Our results highlight the effectiveness of transfer learning as a potential method for effective neural decoding. This advancement opens new possibilities for developing more flexible and data-efficient neural decoding models, with potential applications in augmentative and alternative communication technologies. While our dataset includes only three subjects, prior iEEG studies have shown meaningful results with similar sample sizes~\cite{makin2020machine, willett2023high}. We thus present this as an exploratory study.

\section{Methods}


\begin{figure}[!t]
\centerline{\includegraphics[width=0.5 \textwidth ]{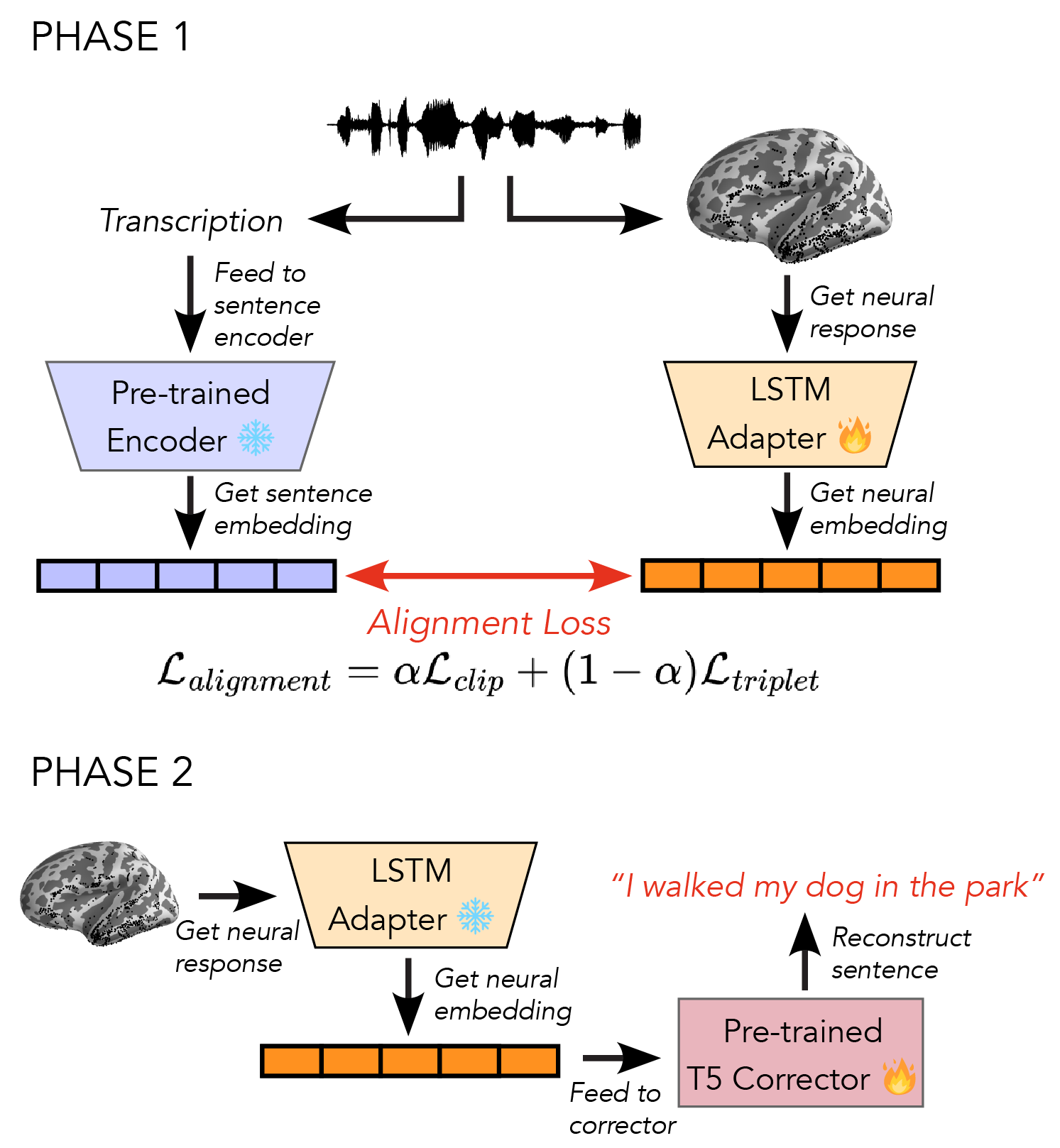}}
\caption{Neuro2Semantic architecture and training methodology. Training is split into 2 phases. In Phase 1, an adapter module is trained to output a neural embedding that is aligned with a fixed sentence embedding. In Phase 2, a corrector module is trained to read out the neural embedding as continuous language.}
\label{fig_methods}
\end{figure}

\begin{figure*}[!t]
\centerline{\includegraphics[width=0.95\textwidth]{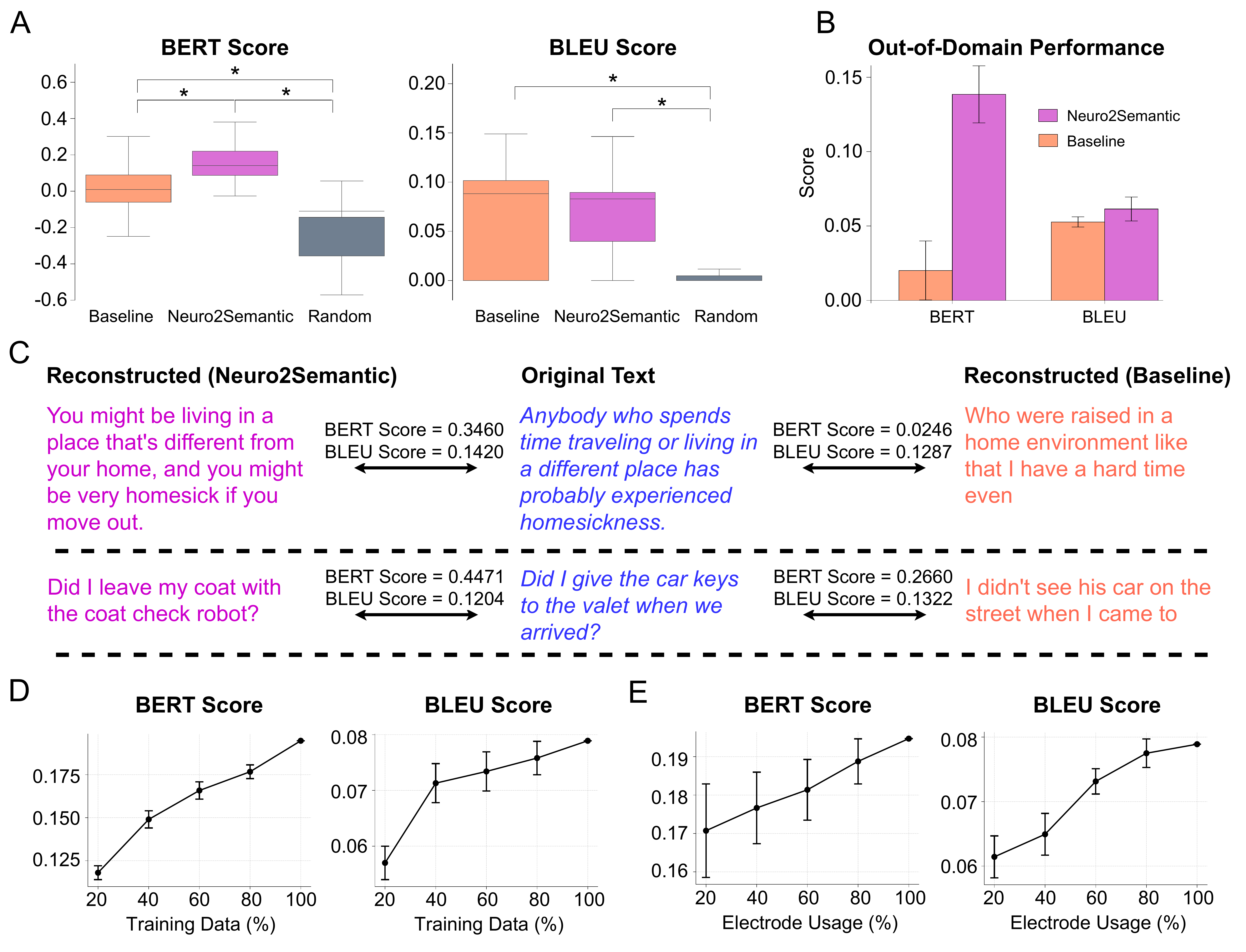}}
\caption{Performance comparison between Neuro2Semantic, baseline, and random control. The BLEU and BERTScore gains correspond to a tangible boost in semantic accuracy
(A) Boxplots of BERTScore (left) and BLEU Score (right) comparing the performance of Neuro2Semantic, the baseline model \cite{tang2023semantic}, and a random control. Significance is indicated with a star based on a paired \emph{t}-test (p $<$ 0.05).
(B) Out-of-domain performance for each method is shown. 
(C) Example sentence reconstructions from Neuro2Semantic (left), original text (middle), and baseline model (right). Samples represent moderately above-average performance rather than extreme cases. (D) The performance of the method is plotted across different percentages of training data. (E) The performance of the method is plotted across different percentages of electrode coverage. All error bars show one standard deviation.
}
\label{fig:results1}
\end{figure*}
\setcounter{footnote}{0}

\subsection{Neuro2Semantic}
The proposed Neuro2Semantic framework, illustrated in Figure \ref{fig_methods}, is designed to map neural signals to their corresponding semantic content through a two-phase training process. In the first phase, an LSTM adapter processes the neural data and aligns it with text embeddings obtained from a pre-trained text embedding model (\textsf{text-embedding-ada-002} \cite{OpenAI2022Embeddings}). In the second phase, we utilize the methodology outlined in the Vec2text framework \cite{morris_text_2023}, which translates these text embeddings back into natural language. During this phase, the LSTM adapter is frozen, and we fine-tune the Vec2text inversion model to reconstruct the text from the aligned neural embeddings. We release our code and trained models.\footnote{Available at \href{https://github.com/SiavashShams/neuro2semantic}{github.com/SiavashShams/neuro2semantic}}

\subsubsection{LSTM adapter and alignment}

Neuro2Semantic employs an LSTM adapter to encode iEEG signals into fixed-dimensional embeddings , thereby aligning them with the semantic space of pre-trained text embeddings. To achieve effective alignment between the neural embeddings generated by the LSTM adapter and the corresponding semantic embeddings, we employ an alignment loss that combines a contrastive objective with a batch-level similarity optimization \cite{radford_learning_2021}. Formally, the adapter is trained using an alignment loss that is a weighted combination of a contrastive loss objective and a triplet margin loss objective:

\begin{equation}
    \mathcal{L}_{\text{alignment}} = \alpha \mathcal{L}_{\text{clip}} + (1 - \alpha) \mathcal{L}_{\text{triplet}}
\label{eq:alignment_loss}
\end{equation}

 This ensures that the neural embeddings are both closely aligned with their corresponding text embeddings and sufficiently distinct from non-corresponding pairs.

\subsubsection{Vec2Text Corrector Module}

The second phase of the Neuro2Semantic framework focuses on transforming the aligned neural embeddings into coherent text sequences. This is accomplished by fine-tuning the Vec2Text corrector module \cite{morris_text_2023}, which is designed to invert text embeddings back into their original textual form. Although the Vec2Text model is pre-trained on large-scale text corpora, fine-tuning it with neural embeddings allows the model to adapt to the specific characteristics of neural embeddings, enhancing its ability to accurately reconstruct the original text from these embeddings.

This inversion task is framed as a controlled generation problem, which aims to generate text \(x\) whose embedding \(\mathbf{\hat{e}}(x)\) closely approximates the target embedding \(\mathbf{e}\).

The model operates iteratively, starting with an initial hypothesis \(x^{(0)}\) and refining it over multiple steps \(t\). At each step, the model minimizes the distance between the current hypothesis embedding \(\mathbf{\hat{e}}(x^{(t)})\) and the target embedding \(\mathbf{e}\), progressively enhancing the coherence and accuracy of the generated text. Mathematically, the goal is to solve the following optimization problem:

\begin{equation}
    \hat{x} = \arg\max_{x} \cos(\mathbf{\hat{e}}(x), \mathbf{e})
\end{equation}

Here, \(\cos(\mathbf{\hat{e}}(x), \mathbf{e})\) represents the cosine similarity between the embedding of the generated text and the target embedding. The optimization seeks to find the text sequence \(x\) that maximizes this similarity.

The iterative refinement process can be expressed as:

\begin{equation}
x^{(t+1)} = \arg\max_{x} p(x | \mathbf{e}, x^{(t)}, \mathbf{\hat{e}}(x^{(t)}))
\end{equation}

where \(p(x | \mathbf{e}, x^{(t)}, \mathbf{\hat{e}}(x^{(t)}))\) is the probability distribution over possible next texts conditioned on the target embedding \(\mathbf{e}\), the current hypothesis \(x^{(t)}\), and its corresponding embedding \(\mathbf{\hat{e}}(x^{(t)})\). The model updates the text hypothesis by comparing the embedding of the current hypothesis \(\mathbf{\hat{e}}(x^{(t)})\) with the target embedding \(\mathbf{e}\), and generating a new text hypothesis that is more aligned with \(\mathbf{e}\).

The Vec2Text model employs an encoder-decoder transformer architecture conditioned on the previous text hypothesis \(x^{(t)}\) and the target embedding \(\mathbf{e}\). This iterative process continues until the cosine similarity \(\cos(\mathbf{\hat{e}}(x), \mathbf{e})\) converges, resulting in text \(x\) that accurately reflects the original semantic and syntactic structure of the text.

During fine-tuning, the LSTM adapter is kept frozen to preserve the semantic alignment established in the first phase. Only the parameters of the Vec2Text corrector module are updated. The process begins by passing the preprocessed iEEG segments through the LSTM adapter to generate fixed-dimensional neural embeddings \(\mathbf{e}_n\). These embeddings, now aligned with the text embedding space, serve as input to the Vec2Text corrector, which aims to reconstruct the original text sequences \(x = (x_1, x_2, \dots, x_T)\), using a standard NLL loss objective.

\subsection{Intracranial Recordings and Data Processing}
Three subjects undergoing surgical evaluation for drug-resistant epilepsy participated. Stereotactic EEG electrodes were implanted intracranially (iEEG) for epileptogenic localization, only responsive electrodes were used (same selection criteria as \cite{mischler2024contextual}). Prior to electrode implantation, all subjects provided written informed consent for research participation. The subjects listened to naturalistic recordings of people engaging in podcast-like conversations between several speakers. There were six different conversations used. In total, the task included about $30$ minutes of speech. The envelope of the high-gamma band ($70-150$ Hz) of the neural recordings during listening was computed using the Hilbert transform \cite{edwards2009comparison} and downsampled to $100$ Hz. A total of 864 electrodes were used across the three subjects after filtering and significance selection. The research protocol was approved by the institutional review board at North Shore University Hospital.

\subsection{Baseline Model}

For baseline comparison, we use the Bayesian decoding method proposed by Tang et al. \cite{tang2023semantic} to generate decoded stimuli. In brief, the method employs beam search to generate candidate continuations, which are then evaluated and ranked using encoding models following Mischler et al. \cite{mischler2024contextual} by modeling the likelihood \(p(R|S)\) of observing brain responses \(R\) given a stimulus \(S\) as a multivariate Gaussian with mean \(\mu = \hat{R}(S)\) and covariance \(\Sigma\) estimated from encoding residuals. We modified this approach for iEEG by using encoding models based on the high-gamma band and by applying fewer and shorter finite impulse response delays to account for the absence of a delayed hemodynamic response. This method was chosen as it represents state-of-the-art results in fMRI decoding and aligns closely with our objective of reconstructing perceived speech semantics through continuous generative decoding.

\section{Experiments and Results}

\subsection{Experimental Setup}
\textbf{Training Procedure}. We trained the model using a leave-one-out cross-validation approach, where the last trial of each story was left out for testing. Each trial was split into sentences, with the corresponding neural data segment from when the sentence was spoken used for training. This setup prevented any anti-causal leakage of information when fine-tuning the language models while allowing the model to train on the semantic content of past sentences within the same conversation. This process was repeated for each of the six stories, with model performance evaluated after each epoch using cross-validation. The held-out trial from each story served as the test set for that split.

During Phase 1, the LSTM adapter was trained for 100 epochs with a batch size of 8, using the Adam optimizer \cite{adam} with a learning rate of 1.3e-3. Once the adapter training was complete, its parameters were frozen for Phase 2, where the pre-trained corrector was fine-tuned for 2 epochs. In this phase, the corrector used only one step for the refinement process.

The CLIP-based contrastive loss was employed with a temperature parameter of $\tau = 0.1$, and the $\alpha = 0.25 $ term was used to control the contribution between the contrastive loss and triplet margin loss. The chosen parameters were optimized by coordinate descent. To evaluate the quality of the reconstructed text, we used two metrics commonly applied in neural decoding analysis, specifically BLEU \cite{papineni2002bleu}, and BERTScore \cite{zhang_bertscore_2020}. These metrics are used to measure both the surface-level (BLEU) and semantic accuracy (BERTScore) of the generated text compared to the ground truth.

\subsection{Performance Comparison: Neuro2Semantic, Baseline, and Phases of Training}

We evaluate Neuro2Semantic against a baseline model from previous work \cite{tang2023semantic} and a random control to rigorously assess our approach's effectiveness. Results are k-fold cross validated using each of the six stories as a test set per fold. Figure \ref{fig:results1}A presents boxplots illustrating the distribution of performance metrics across all sentence pairs, providing a comprehensive view of each model's variability and consistency. Additionally, Figure \ref{fig:results1}C demonstrates sample reconstructed sentences from each model alongside the original ground truth transcriptions, highlighting the qualitative improvements achieved by Neuro2Semantic.

Our Neuro2Semantic model significantly outperforms the baseline, particularly in BERTScore, indicating its suitability for low-data settings. Ablation results comparing the impact of each stage of the training process to the full model and the baseline are shown in Table \ref{neuro2semantic_vs_baseline}.

\begin{table}[!t]
\centering
\caption{Ablation study demonstrating the impact of the two training phases of Neuro2Semantic. Metrics are reported as mean ± standard deviation.  Significant improvements over random are marked with * (p $<$ 0.05) based on a paired \emph{t}-test.}
\renewcommand{\arraystretch}{1.5}
\resizebox{1.0 \columnwidth}{!}{
\begin{tabular}{l|c|c}
\toprule
\textbf{Model} & \textbf{BERT} $\uparrow$ & \textbf{BLEU} $\uparrow$ \\
\midrule
Random & -0.245 ± 0.132 & 0.002 ± 0.003  \\
Baseline - (Tang et al.)  & 0.032 ± 0.127 * & 0.064 ± 0.053 *\\
\midrule
\textbf{Neuro2Semantic (Full Model)} & \textbf{0.195 ± 0.128} * & \textbf{0.079 ± 0.062} * \\
Neuro2Semantic - Adapter Only (Phase 1) & 0.056 ± 0.086 * & 0.068 ± 0.038 *  \\
Neuro2Semantic - Corrector Only (Phase 2) & 0.100 ± 0.099 * & 0.035 ± 0.045 *  \\
\bottomrule
\end{tabular}

}
\label{neuro2semantic_vs_baseline}
\end{table}

\subsection{Zero-Shot Out-of-Domain Performance}

While the previous evaluations assessed Neuro2Semantic's performance in settings where the model encountered familiar semantic contexts, it is also important to evaluate how well the model generalizes to completely unseen semantics. Here, we explore the zero-shot out-of-domain performance by holding out entire stories that the model has not been exposed to during training. This provides a more challenging test of the model's robustness and its ability to reconstruct coherent semantic content when faced with new contexts.

Figure \ref{fig:results1}B presents a bar plot showing Neuro2Semantic’s BERT and BLEU scores which consistently outperform the baseline model. In particular, the BERTScore shows a substantial improvement, suggesting that the model can maintain semantic coherence even when exposed to entirely new stories. This result further indicates that Neuro2Semantic captures broader semantic patterns instead of just memorizing training examples.

\subsection{Impact of Data and Electrode Scaling on Model Performance}

We evaluate how scaling both the training data and the number of electrodes affects the performance of the Neuro2Semantic model. First, we assess the impact of scaling the training data by training the model on random subsets of 20\%, 40\%, 60\%, 80\%, and 100\% of the available data. For each subset percentage, five independent runs were conducted, with the standard deviation across runs displayed as error bars in Figure \ref{fig:results1}D. As the training data increases, we observe significant performance improvements that appear  linear across BERT and BLEU scores. This demonstrates that larger datasets enhance the model’s ability to generalize, leading to more accurate text reconstruction. This emphasizes the scaling potential of our method when exposed to larger datasets.

Similarly, we investigated the effect of varying electrode usage by training the model on a random subset of 20\%, 40\%, 60\%, 80\%, and 100\% of the available electrodes. We ran the experiment five times with a different selected subset for each percentage. The results are presented in Figure \ref{fig:results1}E. We observe similar linear scaling with electrode count, which suggests that Neuro2Semantic could benefit substantially from denser cortical coverage. However, the relatively large error bars imply that some electrodes are substantially more valuable than others for decoding. This suggests that in decoding applications, there are optimal coverage patterns to extract maximally useful information with a fixed quantity of electrodes. 

\section{Discussion}

Neuro2Semantic demonstrates significant advances in neural language decoding through its novel two-phase architecture and efficient data utilization. Unlike classification-based approaches \cite{makin2020machine} or retrieval-oriented frameworks \cite{defossez2022decoding}, our model directly aligns iEEG signals with semantic embeddings, enabling unconstrained text generation without predefined vocabularies. When compared to a replicated current state-of-the-art continuous decoding method \cite{tang2023semantic}, our approach achieves substantially higher semantic accuracy while requiring only 30 minutes of training data, a fraction of the 16+ hours typically needed by existing approaches \cite{tang2023semantic, defossez2022decoding, wang2022open}. Our ablation studies confirm that the initial alignment phase is crucial for performance, rather than merely relying on language model capabilities. This methodology enables zero-shot generalization to unseen semantic content without domain-specific fine-tuning, distinguishing it from previous methods that are constrained by their training vocabularies. Furthermore, our scaling experiments demonstrate consistent performance improvements with increased data and electrode coverage, suggesting significant headroom for enhancement as more data becomes available.

The small sample size and clinical population limit immediate generalizability. Our current goal is to validate feasibility, not yet draw population-level conclusions.
Additionally, as we gather more data, we aim to investigate transformer-based architectures for the alignment phase, which typically require larger datasets but could offer enhanced modeling capacity. These developments would further strengthen Neuro2Semantic's capabilities across different subjects and linguistic contexts.

\section{Conclusion}
We introduce Neuro2Semantic, a transfer learning framework that decodes continuous language from neural signals using pre-trained text embeddings. With just 30 minutes of data, it outperforms existing methods and demonstrates strong zero-shot generalization and unconstrained text generation. The approach scales with larger datasets and limited brain coverage, highlighting its promise for real-world brain-computer interface applications and assistive communication technologies.

\bibliographystyle{IEEEtran}
\bibliography{iclr2025_conference}

\begin{thebibliography}{10}
\providecommand{\url}[1]{#1}
\csname url@samestyle\endcsname
\providecommand{\newblock}{\relax}
\providecommand{\bibinfo}[2]{#2}
\providecommand{\BIBentrySTDinterwordspacing}{\spaceskip=0pt\relax}
\providecommand{\BIBentryALTinterwordstretchfactor}{4}
\providecommand{\BIBentryALTinterwordspacing}{\spaceskip=\fontdimen2\font plus
\BIBentryALTinterwordstretchfactor\fontdimen3\font minus \fontdimen4\font\relax}
\providecommand{\BIBforeignlanguage}[2]{{%
\expandafter\ifx\csname l@#1\endcsname\relax
\typeout{** WARNING: IEEEtran.bst: No hyphenation pattern has been}%
\typeout{** loaded for the language `#1'. Using the pattern for}%
\typeout{** the default language instead.}%
\else
\language=\csname l@#1\endcsname
\fi
#2}}
\providecommand{\BIBdecl}{\relax}
\BIBdecl

\bibitem{kriegeskorte2019interpreting}
N.~Kriegeskorte and P.~K. Douglas, ``Interpreting encoding and decoding models,'' \emph{Current opinion in neurobiology}, vol.~55, pp. 167--179, 2019.

\bibitem{chakrabarti2015progress}
S.~Chakrabarti, H.~M. Sandberg, J.~S. Brumberg, and D.~J. Krusienski, ``Progress in speech decoding from the electrocorticogram,'' \emph{Biomedical Engineering Letters}, vol.~5, pp. 10--21, 2015.

\bibitem{akbari2019towards}
H.~Akbari, B.~Khalighinejad, J.~L. Herrero, A.~D. Mehta, and N.~Mesgarani, ``Towards reconstructing intelligible speech from the human auditory cortex,'' \emph{Scientific reports}, vol.~9, no.~1, p. 874, 2019.

\bibitem{wang2023brainbert}
C.~Wang, V.~Subramaniam, A.~U. Yaari, G.~Kreiman, B.~Katz, I.~Cases, and A.~Barbu, ``Brainbert: Self-supervised representation learning for intracranial recordings,'' \emph{arXiv preprint arXiv:2302.14367}, 2023.

\bibitem{naselaris2011encoding}
T.~Naselaris, K.~N. Kay, S.~Nishimoto, and J.~L. Gallant, ``Encoding and decoding in fmri,'' \emph{Neuroimage}, vol.~56, no.~2, pp. 400--410, 2011.

\bibitem{tang2023semantic}
J.~Tang, A.~LeBel, S.~Jain, and A.~G. Huth, ``Semantic reconstruction of continuous language from non-invasive brain recordings,'' \emph{Nature Neuroscience}, pp. 1--9, 2023.

\bibitem{defossez2022decoding}
A.~D{\'e}fossez, C.~Caucheteux, J.~Rapin, O.~Kabeli, and J.-R. King, ``Decoding speech from non-invasive brain recordings,'' \emph{arXiv preprint arXiv:2208.12266}, 2022.

\bibitem{wang2024semantic}
B.~Wang, X.~Xu, L.~Zhang, B.~Xiao, X.~Wu, and J.~Chen, ``Semantic reconstruction of continuous language from meg signals,'' in \emph{ICASSP 2024-2024 IEEE International Conference on Acoustics, Speech and Signal Processing (ICASSP)}.\hskip 1em plus 0.5em minus 0.4em\relax IEEE, 2024, pp. 2190--2194.

\bibitem{wang2022open}
Z.~Wang and H.~Ji, ``Open vocabulary electroencephalography-to-text decoding and zero-shot sentiment classification,'' in \emph{Proceedings of the AAAI Conference on Artificial Intelligence}, vol.~36, no.~5, 2022, pp. 5350--5358.

\bibitem{liu2024eeg2text}
H.~Liu, D.~Hajialigol, B.~Antony, A.~Han, and X.~Wang, ``Eeg2text: Open vocabulary eeg-to-text decoding with eeg pre-training and multi-view transformer,'' \emph{arXiv preprint arXiv:2405.02165}, 2024.

\bibitem{li2024neural2speech}
J.~Li, C.~Guo, L.~Fu, L.~Fan, E.~F. Chang, and Y.~Li, ``Neural2speech: A transfer learning framework for neural-driven speech reconstruction,'' in \emph{ICASSP 2024-2024 IEEE International Conference on Acoustics, Speech and Signal Processing (ICASSP)}.\hskip 1em plus 0.5em minus 0.4em\relax IEEE, 2024, pp. 2200--2204.

\bibitem{lee24c_interspeech}
J.~Lee, A.~Kommineni, T.~Feng, K.~Avramidis, X.~Shi, S.~R. Kadiri, and S.~Narayanan, ``Toward fully-end-to-end listened speech decoding from eeg signals,'' in \emph{Interspeech 2024}, 2024, pp. 1500--1504.

\bibitem{robinson2016noninvasive}
N.~Robinson and A.~Vinod, ``Noninvasive brain-computer interface: decoding arm movement kinematics and motor control,'' \emph{IEEE Systems, Man, and Cybernetics Magazine}, vol.~2, no.~4, pp. 4--16, 2016.

\bibitem{pandarinath2017high}
C.~Pandarinath, P.~Nuyujukian, C.~H. Blabe, B.~L. Sorice, J.~Saab, F.~R. Willett, L.~R. Hochberg, K.~V. Shenoy, and J.~M. Henderson, ``High performance communication by people with paralysis using an intracortical brain-computer interface,'' \emph{elife}, vol.~6, p. e18554, 2017.

\bibitem{zou2023generalized}
X.~Zou, Z.-Y. Dou, J.~Yang, Z.~Gan, L.~Li, C.~Li, X.~Dai, H.~Behl, J.~Wang, L.~Yuan \emph{et~al.}, ``Generalized decoding for pixel, image, and language,'' in \emph{Proceedings of the IEEE/CVF Conference on Computer Vision and Pattern Recognition}, 2023, pp. 15\,116--15\,127.

\bibitem{xia2024dream}
W.~Xia, R.~de~Charette, C.~Oztireli, and J.-H. Xue, ``Dream: Visual decoding from reversing human visual system,'' in \emph{Proceedings of the IEEE/CVF Winter Conference on Applications of Computer Vision}, 2024, pp. 8226--8235.

\bibitem{benchetrit2024brain}
Y.~Benchetrit, H.~Banville, and J.-R. King, ``Brain decoding: toward real-time reconstruction of visual perception,'' in \emph{The Twelfth International Conference on Learning Representations}, 2024.

\bibitem{willett2023high}
F.~R. Willett, E.~M. Kunz, C.~Fan, D.~T. Avansino, G.~H. Wilson, E.~Y. Choi, F.~Kamdar, M.~F. Glasser, L.~R. Hochberg, S.~Druckmann \emph{et~al.}, ``A high-performance speech neuroprosthesis,'' \emph{Nature}, vol. 620, no. 7976, pp. 1031--1036, 2023.

\bibitem{metzger2023high}
S.~L. Metzger, K.~T. Littlejohn, A.~B. Silva, D.~A. Moses, M.~P. Seaton, R.~Wang, M.~E. Dougherty, J.~R. Liu, P.~Wu, M.~A. Berger \emph{et~al.}, ``A high-performance neuroprosthesis for speech decoding and avatar control,'' \emph{Nature}, vol. 620, no. 7976, pp. 1037--1046, 2023.

\bibitem{huth2016natural}
A.~G. Huth, W.~A. De~Heer, T.~L. Griffiths, F.~E. Theunissen, and J.~L. Gallant, ``Natural speech reveals the semantic maps that tile human cerebral cortex,'' \emph{Nature}, vol. 532, no. 7600, pp. 453--458, 2016.

\bibitem{rybavr2022neural}
M.~Ryb{\'a}{\v{r}} and I.~Daly, ``Neural decoding of semantic concepts: A systematic literature review,'' \emph{Journal of Neural Engineering}, vol.~19, no.~2, p. 021002, 2022.

\bibitem{dash2020decoding}
D.~Dash, P.~Ferrari, and J.~Wang, ``Decoding imagined and spoken phrases from non-invasive neural (meg) signals,'' \emph{Frontiers in neuroscience}, vol.~14, p. 290, 2020.

\bibitem{makin2020machine}
J.~G. Makin, D.~A. Moses, and E.~F. Chang, ``Machine translation of cortical activity to text with an encoder--decoder framework,'' \emph{Nature neuroscience}, vol.~23, no.~4, pp. 575--582, 2020.

\bibitem{10.1162/neco.1997.9.8.1735}
S.~Hochreiter and J.~Schmidhuber, ``Long short-term memory,'' \emph{Neural Comput.}, vol.~9, no.~8, p. 1735–1780, Nov. 1997.

\bibitem{raffel_exploring_2020}
C.~Raffel, N.~Shazeer, A.~Roberts, K.~Lee, S.~Narang, M.~Matena, Y.~Zhou, W.~Li, and P.~J. Liu, ``Exploring the {Limits} of {Transfer} {Learning} with a {Unified} {Text}-to-{Text} {Transformer},'' \emph{Journal of Machine Learning Research}, vol.~21, no. 140, pp. 1--67, 2020.

\bibitem{OpenAI2022Embeddings}
OpenAI, ``Openai api: Embeddings,'' \url{https://platform.openai.com/docs/guides/embeddings}, 2022.

\bibitem{morris_text_2023}
J.~Morris, V.~Kuleshov, V.~Shmatikov, and A.~Rush, ``Text {Embeddings} {Reveal} ({Almost}) {As} {Much} {As} {Text},'' in \emph{Proceedings of the 2023 {Conference} on {Empirical} {Methods} in {Natural} {Language} {Processing}}, H.~Bouamor, J.~Pino, and K.~Bali, Eds.\hskip 1em plus 0.5em minus 0.4em\relax Singapore: Association for Computational Linguistics, Dec. 2023, pp. 12\,448--12\,460.

\bibitem{radford_learning_2021}
A.~Radford, J.~W. Kim, C.~Hallacy, A.~Ramesh, G.~Goh, S.~Agarwal, G.~Sastry, A.~Askell, P.~Mishkin, J.~Clark, G.~Krueger, and I.~Sutskever, ``\BIBforeignlanguage{en}{Learning {Transferable} {Visual} {Models} {From} {Natural} {Language} {Supervision}},'' in \emph{\BIBforeignlanguage{en}{Proceedings of the 38th {International} {Conference} on {Machine} {Learning}}}.\hskip 1em plus 0.5em minus 0.4em\relax PMLR, Jul. 2021, pp. 8748--8763, iSSN: 2640-3498.

\bibitem{mischler2024contextual}
G.~Mischler, Y.~A. Li, S.~Bickel, A.~D. Mehta, and N.~Mesgarani, ``Contextual feature extraction hierarchies converge in large language models and the brain,'' \emph{Nature Machine Intelligence}, pp. 1--11, 2024.

\bibitem{edwards2009comparison}
E.~Edwards, M.~Soltani, W.~Kim, S.~S. Dalal, S.~S. Nagarajan, M.~S. Berger, and R.~T. Knight, ``Comparison of time--frequency responses and the event-related potential to auditory speech stimuli in human cortex,'' \emph{Journal of neurophysiology}, vol. 102, no.~1, pp. 377--386, 2009.

\bibitem{adam}
D.~P. Kingma and J.~Ba, ``Adam: {A} method for stochastic optimization,'' in \emph{3rd International Conference on Learning Representations, {ICLR} 2015, San Diego, CA, USA, May 7-9, 2015, Conference Track Proceedings}, Y.~Bengio and Y.~LeCun, Eds., 2015.

\bibitem{papineni2002bleu}
K.~Papineni, S.~Roukos, T.~Ward, and W.-J. Zhu, ``Bleu: a method for automatic evaluation of machine translation,'' in \emph{Proceedings of the 40th annual meeting of the Association for Computational Linguistics}, 2002, pp. 311--318.

\bibitem{zhang_bertscore_2020}
T.~Zhang, V.~Kishore, F.~Wu, K.~Q. Weinberger, and Y.~Artzi, ``\BIBforeignlanguage{en}{{BERTScore}: {Evaluating} {Text} {Generation} with {BERT}},'' Apr. 2020.

\end{thebibliography}

\end{document}